# Strategic Data Augmentation with CTGAN for Smart Manufacturing: Enhancing Machine Learning Predictions of Paper Breaks in Pulp-and-Paper Production

Hamed Khosravi[a, *], Sarah Farhadpour[b], Manikanta Grandhi[a], Ahmed Shoyeb Raihan[a], Srinjoy Das[c], Imtiaz Ahmed[a]

[a]Department of Industrial & Management Systems Engineering, West Virginia University, Morgantown, WV 26506, USA

[b]Department of Geology and Geography, West Virginia University, Morgantown, WV 26506, USA

[c]School of Mathematical and Data Sciences, West Virginia University, Morgantown, WV 26506, USA

* Corresponding author, *E-mail address:* hamed.khosravi@mail.wvu.edu

**Abstract**

A significant challenge for predictive maintenance in the pulp-and-paper industry is the infrequency of paper breaks during the production process. In this article, operational data is analyzed from a paper manufacturing machine in which paper breaks are relatively rare but have a high economic impact. Utilizing a dataset comprising 18,398 instances derived from a quality assurance protocol, we address the scarcity of break events (124 cases) that pose a challenge for machine learning predictive models. With the help of Conditional Generative Adversarial Networks (CTGAN) and Synthetic Minority Oversampling Technique (SMOTE), we implement a novel data augmentation framework. This method ensures that the synthetic data mirrors the distribution of the real operational data but also seeks to enhance the performance metrics of predictive modeling. Before and after the data augmentation, we evaluate three different machine learning algorithms-Decision Trees (DT), Random Forest (RF), and Logistic Regression (LR). Utilizing the CTGAN-enhanced dataset, our study achieved significant improvements in predictive maintenance performance metrics. The efficacy of CTGAN in addressing data scarcity was evident, with the models' detection of machine breaks (Class 1) improving by over 30% for Decision Trees, 20% for Random Forest, and nearly 90% for Logistic Regression. With this methodological advancement, this study contributes to industrial quality control and maintenance scheduling by addressing rare event prediction in manufacturing processes.

## 1. Introduction

In the realm of Industry 4.0 the development of data-driven modeling techniques, advancements in human-machine interaction, and the application of intelligent manufacturing on a larger scale than what is currently feasible are considered as key future directions [1]. Currently researchers are actively engaged in exploring key design principles, technology trends, and novel architectural designs, in order to realize a strategic roadmap for manufacturers transitioning to Industry 4.0 [2]. These essential elements for successful implementation of Industry 4.0 technologies within manufacturing companies have been analyzed through several case studies [3]. In the context of the technological aspect, researchers have investigated the applications and advantages of deep learning and machine learning in the context of smart manufacturing [4, 5]. Smart Manufacturing can be defined as the evolution of manufacturing technology facilitated by the advancement of information and communication technology [6].

To enhance efficiency, productivity, and sustainability in manufacturing, intelligent technologies, such as IoT, AI and machine learning, should be integrated [7]. Several applications of machine learning and data mining are implemented in the manufacturing sector, including scheduling, monitoring, quality evaluation, malfunction prognosis, and predictive maintenance [8, 9]. Support Vector Machines , Random Forests, and Artificial Neural Networks are extensively used in predictive maintenance across diverse manufacturing and industrial settings [10]. In spite of some limitations such as the requirement for large datasets and the potential for overfitting in complex models, these methods have been shown to be useful in harnessing the power of data-driven decisions when they are used in the manufacturing industry [11].

The field of pulp-and-paper manufacturing has witnessed significant advancements in recent years, driven by the application of data-driven machine learning techniques [12]. Jauhar et al. [13] presented an integrated framework that addresses the dual objectives of performance measurement and prediction within the pulp and paper industry. Their approach combined Data Envelopment Analysis for relative efficiency assessment and Deep Learning (DL) techniques to predict future industry efficiencies [13]. Kalavathi Devi et al. [14] aimed to develop a real-time machine learning model for predicting and controlling key parameters, with a specific focus on regulating steam pressure during the paper manufacturing process. The study harnessed various machine learning algorithms to model critical parameters such as moisture, caliper, and grammage [14]. Shifting the focus to complex manufacturing processes, Xu et al. embarked on the critical objective of developing a robust failure detection and prediction algorithm tailored to real-world manufacturing settings, with a specific application in detecting sheet breaks in a pulp and paper mill. Their methodology combined gated recurrent units and autoencoders to enhance imbalanced learning for failure prediction [15].

The collection of appropriate data is one of the challenges associated with incorporating machine learning into manufacturing operations [16]. Recognizing these challenges posed by unbalanced and insufficient data, Pereira Parente et al. developed a methodology that harnessed data augmentation through Monte Carlo simulation. This approach effectively expanded and balanced the dataset, mitigating data scarcity issues. The authors utilized geometric distances and clustering analysis to calculate output variables, facilitating the classification of generated data into normal or faulty conditions via nearest neighbors search. The predictive machine learning model, a radial basis function neural network, when trained with augmented data, achieved an impressive classification accuracy of nearly 70%. These findings underscore the effectiveness of data augmentation in improving fault detection and diagnostics within industrial processes, offering valuable insights applicable to the pulp and paper industry and beyond [17]. In another study in the realm of fault detection, Kapp et al. [18] present a comprehensive framework for automated event detection. Their primary objectives include the development of a methodology for identifying recurring events or failure patterns using multivariate time series data and the evaluation of two distinct approaches: one based on heuristic segmentation and clustering, and the other incorporating a collaborative method with a built-in time dependency structure. The findings of their research reveal the successful development of a functional framework capable of detecting recurring patterns in manufacturing environments [18].

In actual applications, it is often difficult to collect adequate defect data. This is particularly true when there is an imbalance between the data available for one defect and the data available for the others [19]. A key requirement is that the machine learning model has high generalization performance which essentially quantifies how the model performs on previously unseen data. In this case synthetic data and data augmentation techniques are highly beneficial [20] and there have been numerous techniques developed over the years to generate additional data from the original dataset to compensate for the imbalance problem [21], and several of these methods have been applied on image classification in different areas [22, 23]. However, there is still a gap using these methods on numerical data in the manufacturing sector.

This study tries to answer the key issue of data limitation for machine learning tasks in the manufacturing industry and in this paper, we provide the following contributions:

- We contribute to the field by implementing and comparing the efficacy of SMOTE and CTGAN, two methods for augmenting data specifically chosen for their ability to create synthetic samples resembling imbalanced manufacturing dataset distributions closely.
- The study addresses a gap in the literature regarding the use of machine learning models for highly specialized industrial problems with imbalanced classes by focusing on the manufacturing sector, especially paper break detection in manufacturing machines.
- The proposed methodology consists of an iterative process of evaluating and enhancing models, primarily using recall as a performance metric. It is particularly important in manufacturing environments, in which the failure to detect an issue (such as a paper break) can result in significant downtime and economic losses.
- We propose a versatile framework for dealing with imbalanced data issues, and validate this approach on a manufacturing dataset, thereby demonstrating its practicality in an industrial setting. Our proposed methodology is highly adaptable and can be used across multiple sectors which are affected by the challenges posed by imbalanced data.

**Nomenclature**
RF     random forest
DT     decision tree
LR     logistic regression
AI     artificial intelligence
CTGAN conditional generative adversarial networks
SMOTE synthetic minority oversampling technique
SHAP     SHapley Additive exPlanations

## 2. Data Understanding

This study focuses on operational data gathered from a pulp-and-paper manufacturing facility that utilizes sophisticated paper manufacturing equipment. There are several meters of space devoted to this piece of equipment, which transforms raw materials into finished paper rolls efficiently. A number of sensors are strategically positioned across the length and breadth of this apparatus to collect a variety of measurements. Input parameters include pulp fiber and chemicals, as well as operating parameters such as the type of cutting blade, vacuum pressure at the couch roll, and the rotational velocity of the rotor. In this system, the paper is continuously processed and wound into reels through a continuous rolling mechanism. In spite of this, paper production can sometimes be disrupted when the paper breaks, requiring a halt to operations. An interruption of production triggers a sequence of actions, including halting production, removing the defective reel, rectifying the problem, and finally restarting the machine. There is a substantial economic burden for the mill as a result of this downtime, sometimes exceeding one hour. Therefore, even a marginal reduction in the frequency of these breaks-by 5%, for example-could result in significant savings. This dataset contains 18,398 instances, which were collected as part of the mill's quality assurance and monitoring protocol. It consists of a binary variable 'y', with '1' indicating a break event, a rarity within the dataset, manifested only 124 times, while the remainder is labeled '0', indicating normal operation. Except for x28, which is categorical, and x61, which is binary, most of the predictors, labeled from x1 to x61, are continuous variables. Predictors have been developed which represent both raw material inputs as well as process variables, and have been centered for ease of analysis. In order to maintain the confidentiality of the data, detailed descriptors for these variables have not been provided [24, 25].

Paper breaks, one of the events of interest in this dataset, occur infrequently, making it a classic example of rare event prediction. A dataset with this scarcity is inherently unbalanced, which can adversely affect the performance metrics of any predictive model, including its precision and recall. It is crucial to address this imbalance and accurately predict break events to increase productivity and reduce costs.

*1.1. Data Exploration*

The purpose of this section is to present data visualizations to facilitate a better understanding of the data. In the displayed visualizations, features x29, x34, and x44 are identified as significant predictors of the binary target variable y following feature selection using the Lasso regularization method. By penalizing the absolute size of the coefficients, this regularization approach is especially effective at reducing the complexity of a model, effectively shrinking less important feature coefficients to zero and promoting the selection of more relevant features. The histograms and scatter plots elucidate distinct patterns and levels of dispersion, with x29 and x34 exhibiting some degree of skewness and x44 showing multimodal characteristics. The distributions suggest potential outliers or clusters within the features that could influence modeling efforts. The scatter plots reveal discernible, though not sharply delineated, groupings concerning the binary target, especially in the x34 vs x44 plot, indicating areas where the target classes are more separable. This is less evident in the x29 vs x34 and x29 vs x44 plots, suggesting that x34 and x44 may have a more complex or non-linear relationship with the target variable. Additionally, the scatter plots demonstrate the relationships between feature pairs, although the overlap in classes indicates that, though these features have significant predictive power, their individual and pairwise distributions do not clearly distinguish between classes. The results indicate that advanced analytical techniques are necessary to model these relationships more effectively. In addition, the selected features, while significant, may need to be considered in conjunction with other model complexity or feature interactions to improve prediction accuracy.

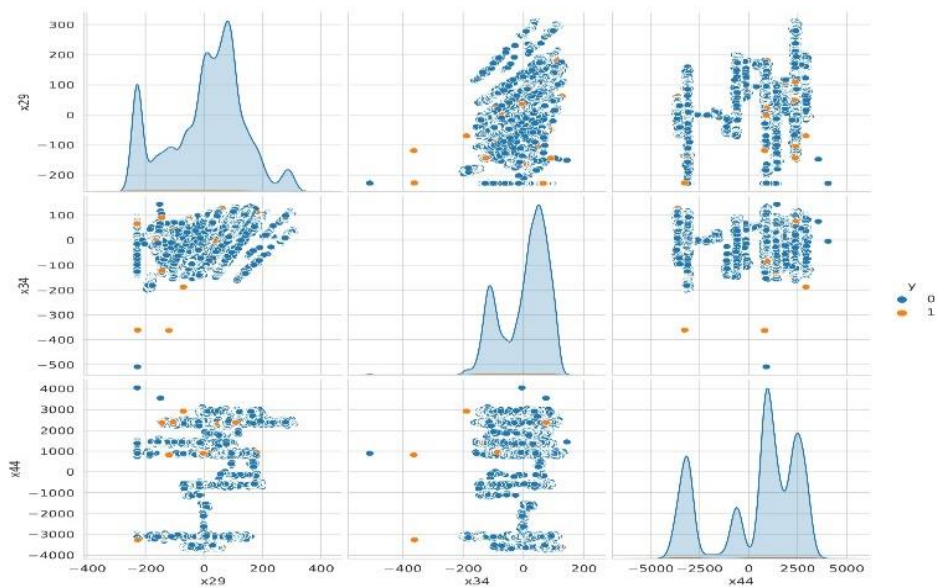

Fig. 1. Pairwise scatter plots and histograms illustrating the relationships between features x29, x34, and x44 for their significance in predicting target

To illustrate the structure within the high-dimensional dataset, t-SNE, a non-linear dimensionality reduction technique, is employed to identify intricate patterns of clustering that linear methods might not be able to reveal. The data points are color-coded in this representation, with green representing class '0' and blue representing class '1'. In the t-SNE plot, it is evident that the data points are not separated into distinct clusters, indicating a degree of class overlap that may complicate the classification process. Class "0" is represented by a large number of green data points, whereas class "1" is represented by sparse and widely scattered blue data points. Due to this disparity, synthetic data generation techniques are required to create additional instances of the minority group.

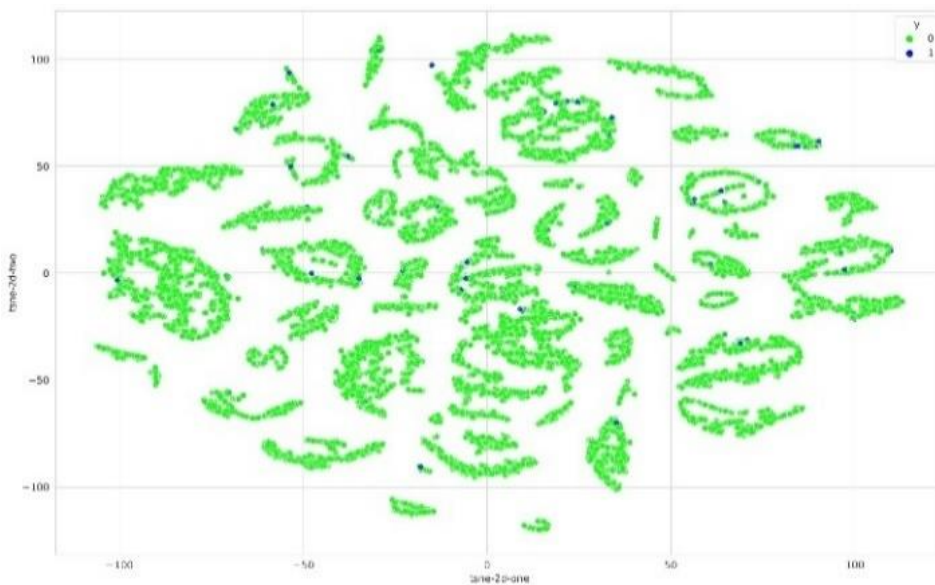

Fig. 2. t-SNE Visualization Highlighting Class Imbalance and Feature Overlap in Selected Feature Space

## 3. Methodology

In this section, first we describe the methods and metrics used in this study. Then, a complete overview of the model utilized is provided.

### 3.1. Terminology

The methods and metrics are as follows:

#### 3.1.1. Decision tree

A Decision Tree (DT) represents a widely used machine learning algorithm capable of addressing both classification and regression tasks. In this structure, every node corresponds to a feature or attribute, each branch represents a decision or rule, and every leaf signifies an outcome, be it categorical or continuous. DTs model human-like reasoning, facilitating intuitive data understanding and insightful interpretations [26]. Their straightforward analysis and accuracy across diverse data types have led to their widespread adoption across various domains [27].

#### 3.1.2. Random Forest

A Random Forest (RF) stands as an ensemble learning strategy used to address classification and regression challenges. Ensemble learning, a machine learning method aimed at enhancing accuracy by combining multiple models to address the same problem, is the overarching concept here [28]. Within the RF algorithm, the initial step entails a random partitioning of the dataset into two segments: one designated for training data to facilitate learning, and the other for validation data to assess the learning's efficacy. Approximately two-thirds of the dataset is allocated to the training data, leaving one-third for validation purposes. Subsequently, numerous decision trees are stochastically generated using "bootstrap samples" derived from the dataset [29]. RFs exhibit diverse tree structures and variable splits, promoting different manifestations of overfitting and accommodating outliers across the various tree models within the ensemble. Consequently, when dealing with classification tasks, the final prediction is determined through a voting mechanism to counteract overfitting, while for regression problems, averaging serves as the solution [28].

*3.1.3. Logistic regression*

Logistic regression stands as a statistical and machine learning model employed in the context of binary classification tasks. It represents a form of regression analysis that is particularly well-suited for predicting a binary outcome based on one or more predictor variables. Contrary to its name, logistic regression is not employed to address regression problems; instead, it serves as a machine learning tool for classification tasks where the dependent variable can take on ~~dichotomous~~ binary values (0/1, -1/1, true/false), while the independent variable can be of a binomial, ordinal, interval, or ratio-level nature [30]. The logistic function, which is an integral component of this model, is expressed as follows:

$$f(x) = \frac{1}{1 + e^{-x}} \quad (1)$$

Where, f(x) represents the outcome derived from a weighted combination of input variables x. If the result exceeds 0.5, it is assigned a value of 1; otherwise, it is assigned a value of 0.

The logistic model originally emerged from the modeling of population growth within the field of ecology [31]. In contemporary times, logistic functions have found application in a wide range of time series prediction problems that extend beyond their ecological origins. Logistic growth is characterized by an initial period of increasing growth, followed by a later stage of decreasing growth as the system approaches a maximum [32].

*3.1.4. Smote*

Synthetic Minority Over-sampling Technique (SMOTE) is a resampling method employed to tackle the challenge of class imbalance within machine learning datasets. Class imbalance arises when one class, typically the majority class, significantly outnumbers another, often referred to as the minority class. This imbalance can result in biased model performance. SMOTE, specifically, is an oversampling strategy designed to address class imbalance by generating synthetic data for the minority class. This approach effectively mitigates the overfitting issue that can occur when using random oversampling [33]. The process of SMOTE involves generating new data points for the minority class by connecting a data point with its k-nearest neighbors. Notably, the synthetic data points created through SMOTE are not direct replicas of the minority class instances. This distinction is crucial to prevent the risk of overfitting during the model training process, as emphasized in reference [34].

*3.1.5. CTGAN*

Conditional Tabular Generative Adversarial Network (CTGAN) represents a generative model specifically crafted to produce synthetic tabular data that upholds the statistical attributes of the source data while enabling the creation of data under specific conditions [35]. Tabular data, which is structured data arranged in a grid format with rows and columns akin to spreadsheets or databases, serves as the basis. It is crucial to underscore that the novel synthetic samples generated by CTGAN do not replicate the instances from the original dataset but instead closely resemble them [36]. This makes it a valuable tool for tasks such as data augmentation, privacy preservation, and generating realistic yet artificial datasets for various data science and machine learning applications.

*3.1.6. Shap*

SHapley Additive exPlanations (SHAP) is a game theory-inspired technique utilized to elucidate the functioning of a machine learning model. Its goal is to create a more interpretable model, and it employs an additive feature attribution method. In this approach, the model's output is defined as a linear sum of its input variables, offering a transparent way to comprehend the model's decision-making process [37]. One of the remarkable aspects of SHAP is that it goes beyond merely assigning average feature importance. Instead, it provides a quantitative explanation for how each feature impacts the model's predictions across the entire dataset. Furthermore, it delves into how each feature specifically influences predictions for individual data points [38]. By leveraging cooperative game theory principles, SHAP provides a robust and intuitive means of achieving model interpretability and transparency in the realm of machine learning.

*3.1.7. Accuracy Assessment*

A confusion matrix is a fundamental tool in the fields of machine learning and statistics, primarily used for assessing the performance of classification models, particularly in binary classification scenarios. It provides a comprehensive summary of the model's predictions in comparison to the actual true outcomes. Typically, this matrix tabulates the number of samples for each combination of the reference class and the predicted class. Although it is common practice for the columns to represent the reference classification labels and the rows to represent the predicted classification labels, it's worth noting that this convention may not always be strictly adhered to [39]. In a binary confusion matrix, the class of interest is often designated as the positive class, while the background class is identified as the negative class (Table 1). In the context of binary classification, there are four possible outcomes to consider. True positives (TP) and true negatives (TN) correspond to the accurate classification of the positive and negative classes, respectively. A false positive (FP) occurs when the background class is incorrectly classified as the positive class, whereas a false negative (FN) arises when the positive class is mistakenly assigned to the background class [40].

Table 1. The confusion matrix for binary classification. TP = True Positive, TN = True Negative, FP = False Positive, and FN = False Negative

|  |  | Reference Data | |
| --- | --- | --- | --- |
|  |  | Positive | Negative |
| **Classification Result** | Positive | TP | FP |
|  | Negative | FN | TN |

From this cross-tabulation, it is possible to calculate several different metrics. Precision (as shown in Equation (2)) signifies the ratio of correctly classified samples among those predicted to be positive. Recall (as defined in Equation (3)) signifies the ratio of correctly classified instances within the positive class relative to the reference data [41]. Overall accuracy (as presented in Equation (4)) denotes the ratio of samples correctly classified (both positive and negative) to the total number of samples.

$$Precision = \frac{TP}{TP + FP} \tag{2}$$

$$Recall = \frac{TP}{TP + FN} \tag{3}$$

$$Overall\ Accuracy = \frac{TP + TN}{TP + TN + FP + FN} \tag{4}$$

*3.2. Applied methodology in the study*

A description of the methodology used in this study is presented in this section. The flowchart provided illustrates this method for addressing challenges associated with imbalanced datasets in machine learning models. Based on this flowchart, we can see how a structured approach is followed from the input of a dataset to the identification of the most effective model. As part of the initial stages, the data must be preprocessed and partitioned into training and test sets. In the process of fitting initial machine learning models, the balance of the dataset is assessed. Iterative applications of data augmentation techniques are advocated if there is an imbalance identified so that the underlying data structures of the generated data closely resemble those of the real data.

An in-depth discussion of this iterative process is provided in the accompanying pseudocode. In each iteration, machine learning models are trained using the current version of the training set. On the test set, the performance metrics are evaluated after model training, and the results are cataloged as "CurrentPerformance". The model's performance is evaluated based on desirable metrics. A data augmentation technique may be used to enhance the training set if it is necessary to improve performance. As a result of iterative refinement with data augmentation, the model can improve its performance in scenarios where the original dataset was imbalanced, resulting in inaccurate predictions.

As part of the final phase of the methodology, after achieving satisfactory model performance, a complete comparison of the machine learning models is conducted. We evaluate the performance metrics across iterations to determine which model is most effective. The model is then designated for deployment or further investigation. In addition to addressing the limitations of imbalanced datasets, this methodology highlights the importance of continuing to refine and evaluate models.

In this study, we demonstrate how machine learning modeling for paper manufacturing can address challenges associated with imbalanced datasets. Considering how important it is to detect all possible breaks in paper manufacturing machines, we primarily evaluate models by recall metrics. For data augmentation, we use both SMOTE and CTGAN. These techniques are applied iteratively so that the augmented data retains the original characteristics of the original manufacturing data. With each iteration, machine learning models - specifically Random Forest (RF), Decision Tree (DT), and Logistic Regression - are trained and assessed based on their performance metrics. After carefully comparing these models, we aim to select the most robust model suitable for manufacturing applications. It is critical to note that this comprehensive approach not only addresses data imbalances but also strives to provide accurate and reliable predictive capabilities tailored for the manufacturing industry.

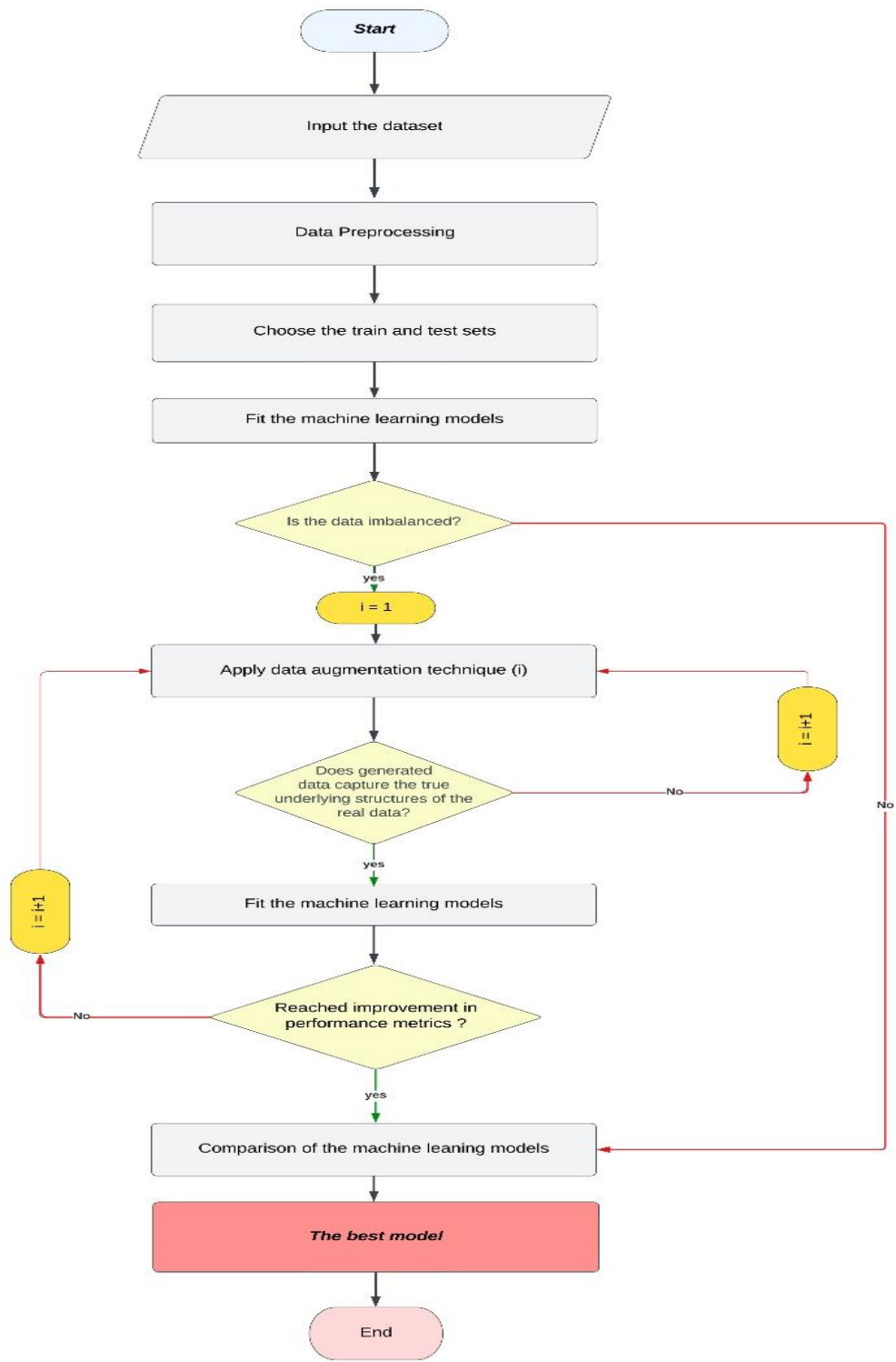

Fig. 3. The flowchart of the iterative Machine Learning model optimization with data augmentation techniques

```
Algorithm 1: Iterative Machine Learning Model Optimization
              with Data Augmentation Techniques
```
**Inputs**: Dataset D; Machine Learning (ML) models; Data Augmentation Techniques T; Maximum Number of Applied Augmentation Techniques n; Performance Metrics M
**Output**: Best ML Model; Trained Dataset; Evaluated Performance Metrics

*// Step 1: Initializing*
01: Input and preprocess dataset D
02: Split D into train_set and test_set
03: Initialize iteration counter: i = 0
04: Initialize BestPerformance = 0
*// Step 2: Model Training and Data Augmentation Loop*
05: **While** $i \leq n$ **do**:
   06: Fit ML models on train_set
   07: CurrentPerformance = Evaluate performance on test_set using M
   08: **IF** CurrentPerformance > BestPerformance **Then**:
      09: BestPerformance = CurrentPerformance
   10: **ELSE IF** data in train_set is imbalanced **Then**:
      11: Apply the ith data augmentation technique from T to train_set
      12: **IF** augmented data retains real data structure **Then**:
         13: Increment i by 1
         14: Continue to next iteration
      15: **ELSE**:
         BREAK
   16: **END IF**
17: **END While**
*// Step 3: Model Comparison*
*18:* Compare performance metrics for all trained ML models
*19:* Select the best model based on the performance metrics
20: **Return** Best ML Model, Trained Dataset, and Performance Metrics
21: *end*

Fig. 4. The methodology Pseudo-code

## 4. Results and Discussions

### 4.1. Real data

The real dataset was used to train and test machine learning models. 70% of the data was used to train models and 30% was used to test them. Table 2 shows 5491 observations that belong to class 0 and 29 observations that belong to class 1.

Table 2. The performance of the models using only the real data

| Model | | DT | RF | LR |
|---|---|---|---|---|
| **Class 0** | Precision | 0.99 | 0.99 | 0.99 |
| | Recall | 0.99 | 0.99 | 0.99 |
| | F1 score | 0.9999 | 0.9999 | 0.9999 |
| **Class 1** | Precision | 0.44 | 0.85 | 0.50 |
| | Recall | 0.55 | 0.59 | 0.07 |
| | F1 score | 0.49 | 0.69 | 0.12 |
| **Overall Accuracy** | | 0.9940 | 0.9973 | 0.9947 |

*4.2. SMOTE-generated data*

In this section, we employ the SMOTE technique to generate artificial data derived from the original dataset. Based on the training data, the SMOTE technique was applied to produce 5,056 new samples. This quantity was selected specifically in order to compensate for existing data imbalances. Thus, a total of 5,151 and 12,783 samples are now included in Classes 0 and 1. It ensures a ratio of 4 to 10 between minor and major classes, resulting in a balanced dataset.

Our first step is to compare the distribution of the real and fake data. This can be performed using various methods which includes kernel based hypothesis testing for high dimensional data [42], comparing various features of the ground truth and generated distributions and measuring classification performance with the synthetic data. Here we adopt the last two approaches and identify the most important features based on Lasso models (x29, x34, x44) as well as categorical features (x28). Figure 5 demonstrates how well SMOTE has generated the fake data by analyzing and comparing it to actual data. Classification performance with the augmented data is discussed later.

Cumulative Distribution Plot and Probability Density Plot for a few selected features are shown in Figure 5. The cumulative distribution plot for 'x28' shows a similar distribution between the real and fake data up to a value of 100. As a result, the real and synthetic categorical counts diverge beyond this point, indicating variations. Both the real and fake data categories exhibit distinct peaks on the density plot for this feature. A potential overrepresentation of certain categories is clear in the synthetic data by the more pronounced and defined peaks around the 80 mark meaning the generated data has not reflected this feature categorically. It is apparent that both the real and fake data distributions for 'x29' almost overlap, indicating that the synthetic data is effective at replicating the real data. However, slight deviations can be observed around the -200 and 200 marks. The feature x29's density plot illustrates pronounced peaks for the real data, while the fake distribution appears smoother. While both distributions share similarities around the zero point, there are some major differences.

In the 'x34' feature, there is a noticeable difference in distribution between real and fake data around the -200 mark, suggesting that the synthetic data may not have adequately replicated certain nuances of the real data. In their density plots, both data sets appear to share a similar trend around the 0 mark. However, the fake data distribution displays a broader peak indicating an oversmoothing process during synthetic data generation. Based on the cumulative plot for 'x44', the real and fake data appear to match fairly closely until around the -1000 mark. However, after this, the fake data appears to fluctuate more significantly. Although both distributions share a peak near zero in their density plots, the real data displays more pronounced peaks at extreme values, which are not as apparent in the fake data.

Next, we conducted principal component analysis (PCA) to determine whether the synthetic data generation process accurately reflects the true underlying structures of the actual data. Figure 6 illustrates the first two components of PCA for the real and fake data generated by SMOTE. PCA is a powerful visualization tool that provides variance captures, spread, and distribution.

As can be seen from the "Real data" plot in Figure 6, the data demonstrates a strong structure or grouping. The data points scattered vertically indicate lesser common patterns or outliers in the original data, whereas the "Fake data" shows multiple structured patterns. The dense clustering near the x-axis indicates the dominant patterns or groups in the original data. Consequently, the synthetic data generation may have emphasized certain patterns or introduced new ones.

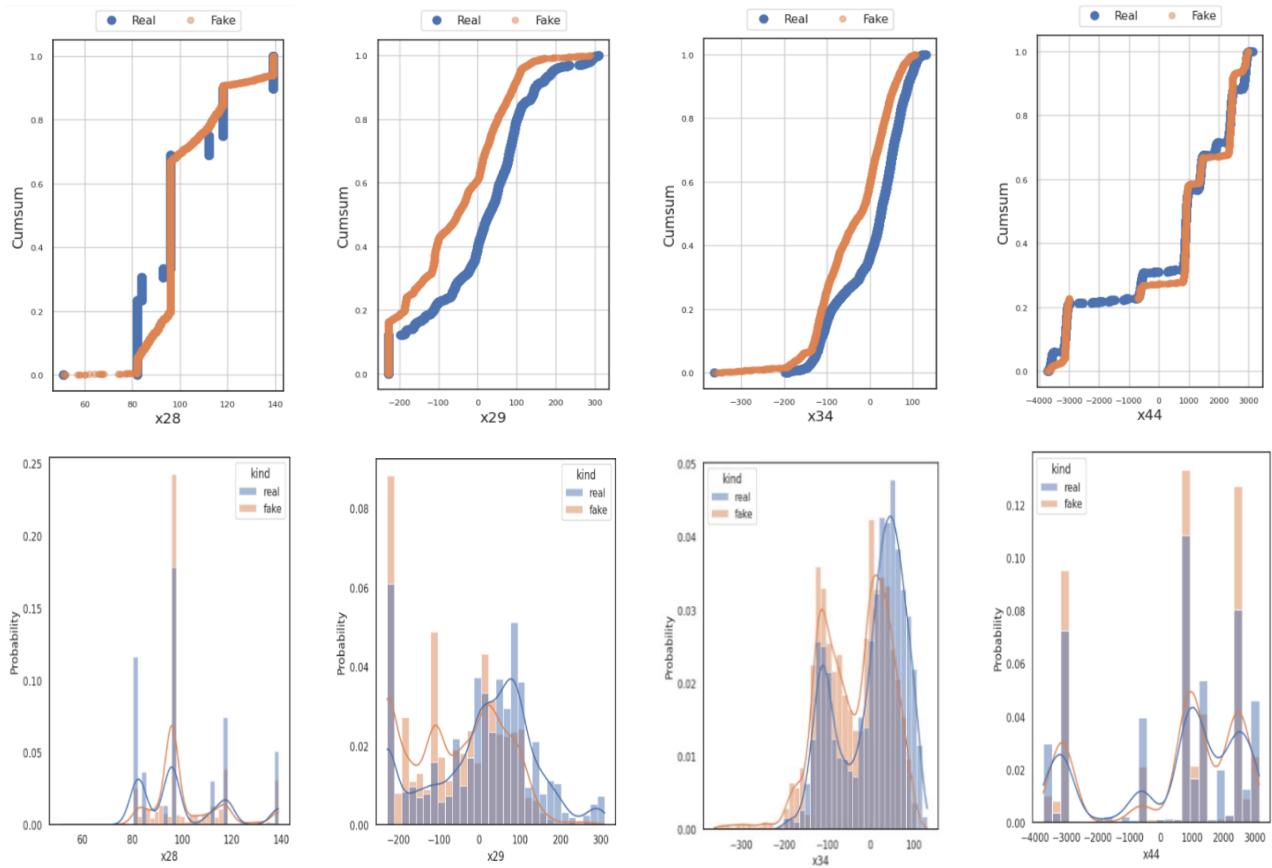

Fig. 5. Comprehensive comparison between the distributions of real and fake (synthetically generated by SMOTE) data across multiple features

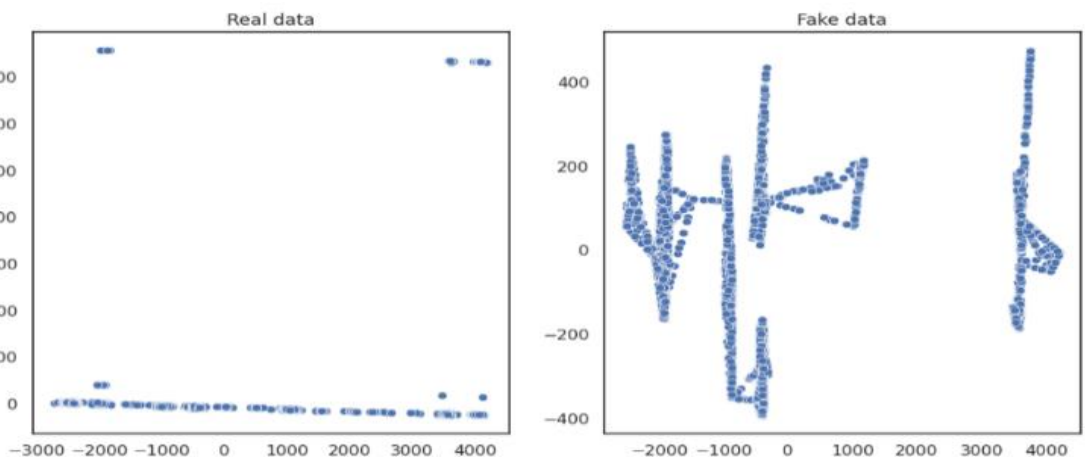

Fig. 6. Comparison between the real and fake (synthetically generated by SMOTE) data considering the first two components of PCA

After checking the quality of the generated data with Smote, we analyze the performance of machine learning models using the combination of smote-generated data and real data. The Figure 7 illustrates the recall performance of three distinct models based on a combination of real and SMOTE data, which were trained 25 times to account for variability and ensure the robustness of the observed results. The Decision Tree demonstrates the inherent variability of the model, emphasizing the model's sensitivity to various data subsets. It is noteworthy that the Decision Tree achieved a high recall value for Class 0. There appears to be a concentration of recall values between 0.6 and 0.7 in Class 1, with a median value of 0.62, which is better than using only the actual data. As a result of its aggregated decision-making approach, the Random Forest model offers a more stable recall rate. While recalls are similar to the Decision Tree, the values for Class 1 are more consistent than those for Class 2. According to logistic regression, recall values for Class 0 are confined to a narrow range between 0.9 and 0.95. Similarly, for Class 1, a majority of the recall values seem to be around the 0.65 mark, with a few values slightly higher or lower. Overall, SMOTE-generated data improved all three models' recall for class 1. In summary, while the synthetic data generated by SMOTE has improved the recall of the models and aligns well with the real data in some instances, there are evident disparities in others (clear in Figures 5 and 6). So, SMOTE has not generated qualified data based on the real data. In the following, we analyze the data generated by CTGAN.

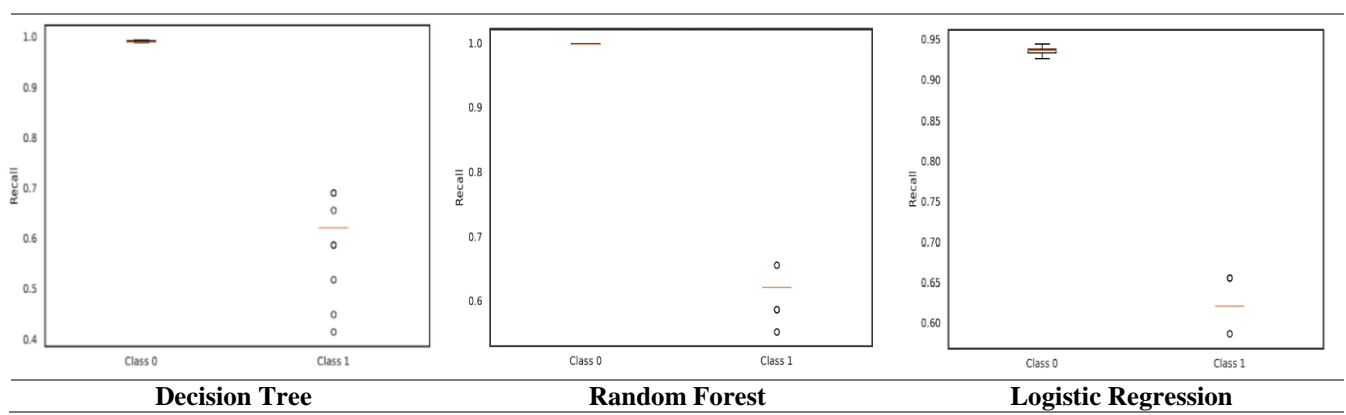

| Decision Tree | Random Forest | Logistic Regression |

Fig.7. The Recall of two classes using the generated data with SMOTE generated and real data for different models

4.3. CTGAN-generated data

CTGAN is employed in this section to generate artificial data from the original dataset. To ensure consistency and validity, the same parameters have been used as in SMOTE. First, we examine the spread and distribution of the CTGAN-generated data compared to real data. As shown in Figure 8, the cumulative distribution plots of feature x28 are strikingly closely aligned between the real and fake datasets, demonstrating the ability of synthetic data to replicate the categorical distribution of real data. In addition, the probability density plots further support this assertion, which is a testament to the effectiveness of the data generation methodology. Overall, the peaks and troughs of the synthetic dataset align
closely, demonstrating that it reflects the categorical distributions underlying the real data, despite minor deviations. The cumulative distribution plots again emphasize the similarity between the real and fake data when it comes to continuous features. As a result, the curves for x29 and x34 are almost congruent, indicating a high degree of similarity in the distribution of values. Further, probability density plots offer a-­‐detailed perspective on data distribution, revealing nuances. Despite minor deviations in density peak shapes and spreads between the two datasets, the overall shape and spread of the x44 feature remain consistent. In light of the results of this study, it is evident that synthetic data can replicate the distributions and relationships inherent in real data in remarkable ways. Next, Figure 9 shows the first two components of PCA for the real and fake data generated by CTGAN, to verify whether the data generation process has captured the underlying structure of the real data.

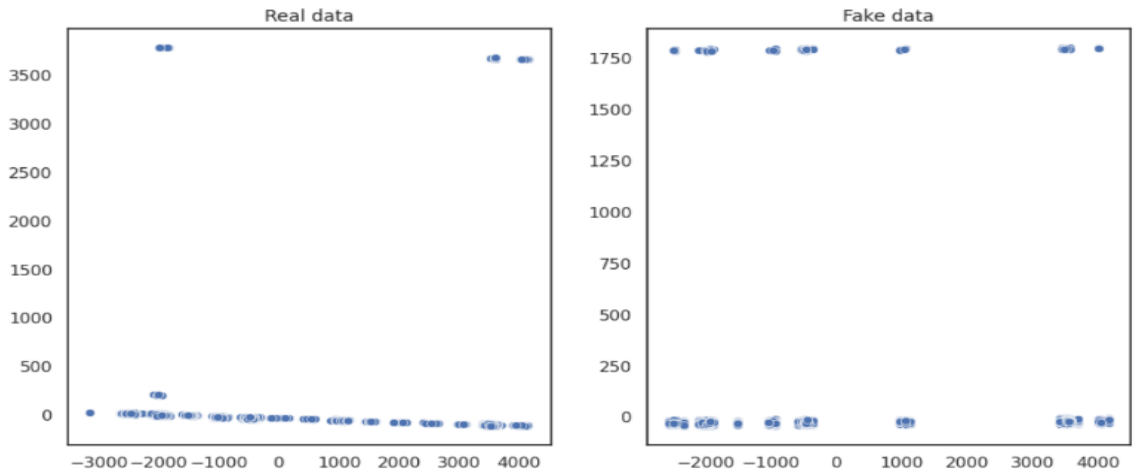

Fig. 8. Comprehensive comparison between the distributions of real and fake (synthetically generated by CTGAN) data across multiple features

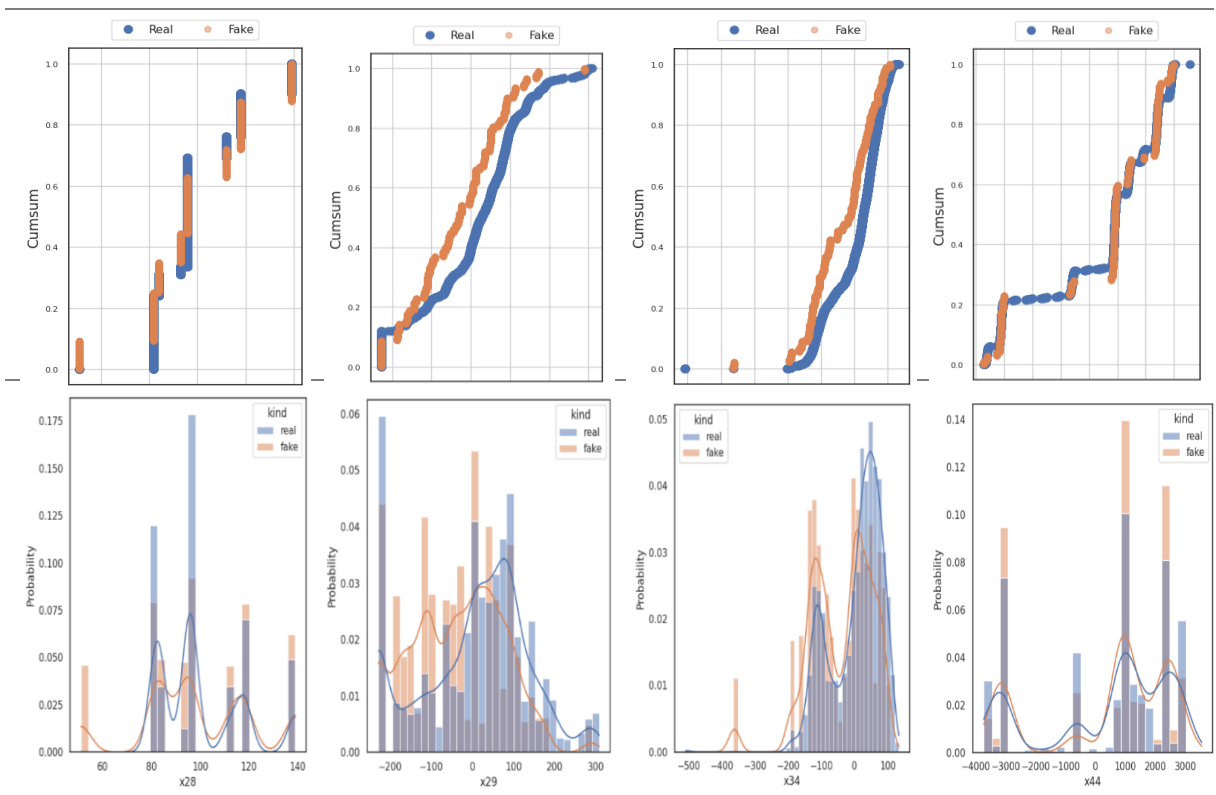

Fig. 9. Comparison between the real and fake (synthetically generated by CTGAN) data considering the first two components of PCA

Figure 9 clearly illustrates that the synthetic data generation process has been highly effective. The fake data reflects the distribution of the real data impressively, especially along the x-axis. As compared with the real data, the fake data exhibits a strikingly similar pattern to the real data, exhibiting concentrated clustering in the lower left quadrant. Both the clustering nuances and the modes of the distributions indicate a close resemblance between the two datasets, highlighting the quality and accuracy of the data generation method. Next, we analyze the machine learning models using the combined data. In Figure 10, it is evident that all three models performed well for Class 0, like real data and the combination of real and SMOTE-generated data. In Class 1, the best Decision Tree model showed an improved recall range from 0.62 to 0.85 with a median recall of 0.73 as compared to the other two cases (real data and combined real and SMOTE data). Even the worst result in this dataset is better than using only real data, it should be noted. Random Forest has the same median as Decision Tree, and it is more consistent. However, the best result does not surpass the Decision Tree. As compared to Class 0, Logistic Regression recall distribution exhibits greater variability, with its median value located around 0.70. There are a few outliers that indicate periodic inconsistencies in the performance of the Logistic Regression model for this class. A robust training foundation appears to be provided when the real and CTGAN-synthesized data are combined, with the decision tree model improving recall by over 30%, the random forest model by over 20%, and the logistic regression model by approximately 90%.

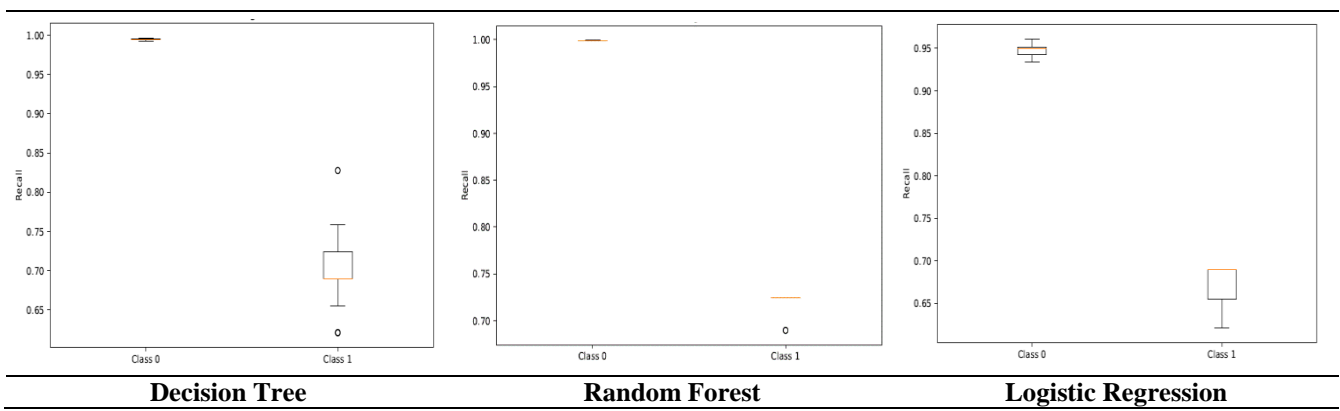

| Decision Tree | Random Forest | Logistic Regression |

Fig.10. The Recall of two classes using the generated data with CTGAN generated and real data for different models

In the following, we compare the performance of the models based on recall in different scenarios for the two classes. With the adoption of data augmentation strategies, specifically CTGAN and SMOTE, it becomes evident that the recall for class 1 exhibits notable improvements when compared across various classifiers. Using CTGAN-processed data, the Decision Tree classifier approaches an optimal recall value, while utilizing SMOTE-enhanced data produces a nearly similar performance. Similarly, Random Forest models exhibit substantial recall improvements when exposed to augmented datasets containing CTGAN and SMOTE, significantly outpacing their real counterparts. While Logistic Regression does not attain exemplary recall levels as its counterparts have, it still shows significant improvements in recall when trained on augmented data. For class 0, it is also noteworthy to note that commendable results are consistently achieved regardless of the model or data technique used. As a result, all three models performed better on CTGAN-generated data combined with real data than the other two cases, with Random Forest showing the most consistent and efficient results among them. As a next step, we will analyze the performance and procedure of this method. This section analyzes the best model, which is the Random Forest training on CTGAN-generated and real data, comparing it with only real data to illustrate the improvement. A comparison of Random Forest classification results on two different datasets is presented in the confusion matrixes presented in Figure 11.

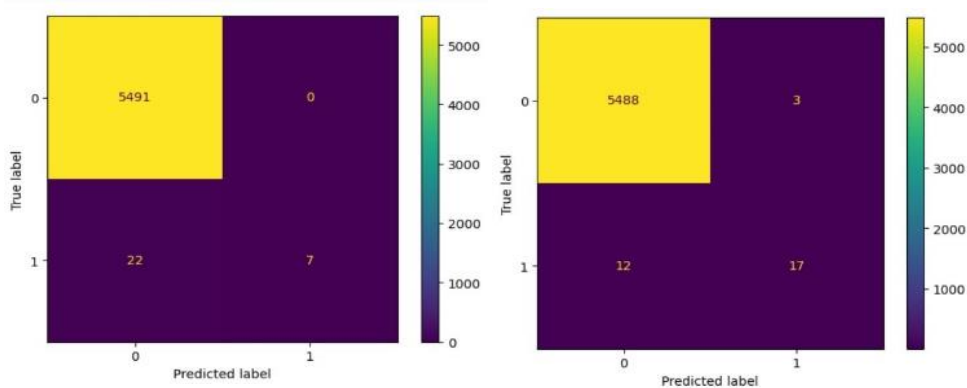

Fig. 11. Comparison of Random Forest Classification Results:
CTGAN-Augmented vs. Real Data

According to the first matrix, RF trained solely on real data yielded 7 true positives and 22 false negatives for Class 1. On the other hand, the second matrix, which is associated with RF trained on real and CTGAN-augmented data, exhibited an increased number of true positives, 17, and a reduced number of false negatives, 12. The number of true positives for Class 0 has also increased and has reached the ideal level of precision.

As a result of integrating CTGAN data augmentation, it is evident that the correct classification of Class 1 instances has been significantly improved, illustrating the value of using synthetic data augmentation techniques to enhance model performance. This model will be examined in more detail in the following. Next, we highlight the main features that are used in the models for each of the classes derived from SHAP.

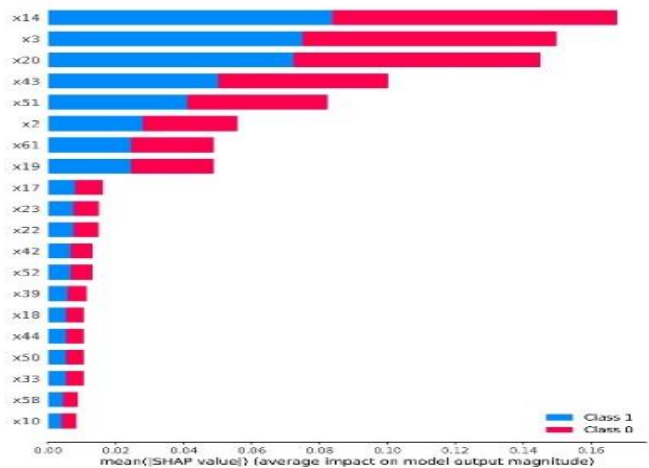

Fig. 12. Comparative SHAP-based Feature Importance for Class 1 and 0

Using SHAP values, feature importance is illustrated in the depicted bar chart, segmented for Class 1 and Class 0. Each bar represents a distinct feature. In particular, features such as x14, x3, x20, and x51 demonstrate considerable relevance to Class 1. In contrast, features such as x10, x58, x50, and x44 demonstrate a greater importance for Class 0. As a result of this distinction, it is evident that the significance of features varies across the two classes, emphasizing the need to interpret model outcomes with nuanced analysis. The subsequent 3D scatter plot provides a detailed visualization of the distribution and separation of these features, particularly x3, x14, and x20 (the most significant features) concerning class boundaries.

According to the 3D scatter plot shown in the figure, features x3, x14, and x20 are intricately related in distinguishing the boundaries for Classes 1 and 0. Based on our observations, x3 ranges from approximately -20 to 15, with Class 0 data densely populated around -15 and Class 1 data concentrated closer to values of 5. In Feature x14, a broad range of values are observed between -1.0 and 2.5; however, Class 1 exhibits a higher concentration between 1.0 and 2.0, whereas Class 0 exhibits a more even distribution. As for the x20 axis, which serves as our z-axis, values oscillate between -1.5 and 3.0. At the extremes, Class 0 is dominant, especially near 2.5, whereas Class 1 dominates the mid-range, around 0.5. These evident clusters and areas of separation are helpful for understanding the complex relationships within the dataset.

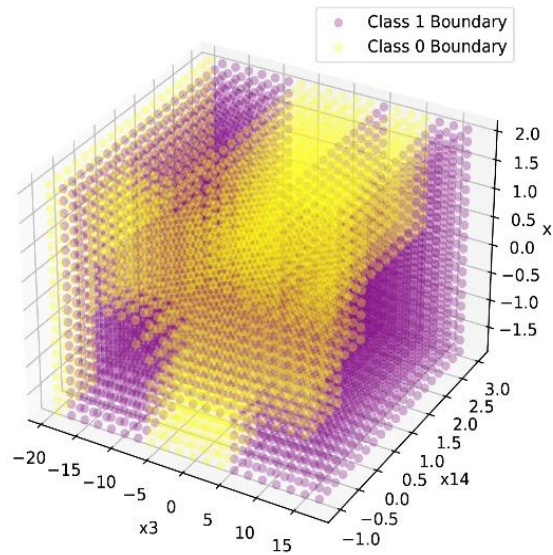

Fig. 13. Three-dimensional distribution of the most influential
features: x3, x14, and x20, demarcating Class 1 and Class 0 boundaries

## 5. Conclusion

In this research we have proposed significant contributions to the field of predictive maintenance within the pulp-and-paper industry, a sector where unscheduled downtime due to paper breaks can have significant economic repercussions. Using a novel data augmentation framework leveraging CTGAN and SMOTE to address the challenge of infrequent break events, our study advances the ability to detect paper breaks, thereby providing an alternative strategy to mitigate substantial production losses and costs associated with paper breaks.

The proposed methodology emphasizes an iterative process of model evaluation and enhancement, focusing on recall as a key performance metric. This is particularly important in the manufacturing industry, where undetected failures have a far greater financial cost than false positives. Based on the augmented datasets, we have observed significant improvements in recall—over 30% for Decision Trees, 20% for Random Forests, and nearly 90% for Logistic Regression. In this case, the rarity of paper breaks highlights how machine learning can be finely tuned to handle imbalanced data in industrial settings.

This study has two practical implications. First, manufacturers can minimize downtime by reducing the likelihood of undetected breaks, which directly leads to improved operational efficiency and cost savings. In addition, the deployment of such predictive models can facilitate maintenance scheduling, decreasing unscheduled stops in both frequency and duration. As a result of this proactive maintenance approach, not only are immediate losses minimized, but the lifespan of manufacturing equipment is also increased, thereby reducing the need to purchase new equipment over an extended period.

In summary, this paper provides tangible benefits to the manufacturing industry in addition to its technical contributions. This study bridges the gap between theoretical machine learning approaches and practical, cost-saving measures in manufacturing processes by demonstrating the value of CTGAN-enhanced datasets in an industrial application.

## Acknowledgements

We would like to express our profound appreciation to the Department of Industrial & Management Systems Engineering (IMSE) for their consistent and invaluable support during this research project. We are also immensely thankful to our colleagues and experts in the area of smart manufacturing. The collaboration, expertise, and constructive insights they provided have greatly enhanced the quality and depth of our study.

## References

[1] R. Y. Zhong, X. Xu, E. Klotz, and S. T. Newman, "Intelligent Manufacturing in the Context of Industry 4.0: A Review," *Engineering,* vol. 3, no. 5, pp. 616-630, 2017/10/01/ 2017, doi: https://doi.org/10.1016/J.ENG.2017.05.015.

[2] M. Ghobakhloo, "The future of manufacturing industry: a strategic roadmap toward Industry 4.0," *Journal of Manufacturing Technology Management,* vol. 29, no. 6, pp. 910-936, 2018, doi: 10.1108/JMTM-02-2018-0057.

[3] R. Pozzi, T. Rossi, and R. Secchi, "Industry 4.0 technologies: critical success factors for implementation and improvements in manufacturing companies," *Production Planning & Control,* vol. 34, no. 2, pp. 139-158, 2023/01/25 2023, doi: 10.1080/09537287.2021.1891481.

[4] J. Wang, Y. Ma, L. Zhang, R. X. Gao, and D. Wu, "Deep learning for smart manufacturing: Methods and applications," *Journal of Manufacturing Systems,* vol. 48, pp. 144-156, 2018/07/01/ 2018, doi: https://doi.org/10.1016/j.jmsy.2018.01.003.

[5] T. Wuest, D. Weimer, C. Irgens, and K.-D. Thoben, "Machine learning in manufacturing: advantages, challenges, and applications," *Production & Manufacturing Research,* vol. 4, no. 1, pp. 23-45, 2016/01/01 2016, doi: 10.1080/21693277.2016.1192517.

[6] B. Wang, F. Tao, X. Fang, C. Liu, Y. Liu, and T. Freiheit, "Smart Manufacturing and Intelligent Manufacturing: A Comparative Review," *Engineering,* vol. 7, no. 6, pp. 738-757, 2021/06/01/ 2021, doi: https://doi.org/10.1016/j.eng.2020.07.017.

[7] K. Haricha, A. Khiat, Y. Issaoui, A. Bahnasse, and H. Ouajji, "Recent Technological Progress to Empower Smart Manufacturing: Review and Potential Guidelines," *IEEE Access,* vol. 11, pp. 77929-77951, 2023, doi: 10.1109/ACCESS.2023.3246029.

[8] A. Dogan and D. Birant, "Machine learning and data mining in manufacturing," *Expert Systems with Applications,* vol. 166, p. 114060, 2021/03/15/ 2021, doi: https://doi.org/10.1016/j.eswa.2020.114060.

[9] M. A. Farahani, M. McCormick, R. Harik, and T. Wuest, *Time-Series Classification in Smart Manufacturing Systems: An Experimental Evaluation of State-of-the-Art Machine Learning Algorithms*. 2023.

[10] Z. M. Çınar, A. Abdussalam Nuhu, Q. Zeeshan, O. Korhan, M. Asmael, and B. Safaei, "Machine Learning in Predictive Maintenance towards Sustainable Smart Manufacturing in Industry 4.0," *Sustainability*, vol. 12, no. 19, doi: 10.3390/su12198211.

[11] I. M. Cavalcante, E. M. Frazzon, F. A. Forcellini, and D. Ivanov, "A supervised machine learning approach to data-driven simulation of resilient supplier selection in digital manufacturing," *International Journal of Information Management,* vol. 49, pp. 86-97, 2019/12/01/ 2019, doi: https://doi.org/10.1016/j.ijinfomgt.2019.03.004.

[12] H. Zhang, J. Li, and M. Hong, "Machine Learning-Based Energy System Model for Tissue Paper Machines," *Processes*, vol. 9, no. 4, doi: 10.3390/pr9040655.

[13] S. Jauhar, P. Raj, S. Kamble, S. Pratap, S. Gupta, and A. Belhadi, "A deep learning-based approach for performance assessment and prediction: A case study of pulp and paper industries," *Annals of Operations Research,* 01/16 2022, doi: 10.1007/s10479-022-04528-3.

[14] T. Kalavathi Devi, E. B. Priyanka, and P. Sakthivel, "Paper quality enhancement and model prediction using machine learning techniques," *Results in Engineering,* vol. 17, p. 100950, 2023/03/01/ 2023, doi: https://doi.org/10.1016/j.rineng.2023.100950.

[15] D. Xu, Z. Zhang, and J. Shi, "Failure Prediction Using Gated Recurrent Unit and Autoencoder in Complex Manufacturing Process," in *2022 4th World Symposium on Artificial Intelligence (WSAI)*, 23-25 June 2022 2022, pp. 68-73, doi: 10.1109/WSAI55384.2022.9836412.

[16] S. Kumar *et al.*, "Machine learning techniques in additive manufacturing: a state of the art review on design, processes and production control," *Journal of Intelligent Manufacturing,* vol. 34, no. 1, pp. 21-55, 2023/01/01 2023, doi: 10.1007/s10845-022-02029-5.

[17] A. Pereira Parente, M. B. de Souza Jr, A. Valdman, and R. O. Mattos Folly, "Data Augmentation Applied to Machine Learning-Based Monitoring of a Pulp and Paper Process," *Processes*, vol. 7, no. 12, doi: 10.3390/pr7120958.

[18] V. Kapp, M. C. May, G. Lanza, and T. Wuest, "Pattern Recognition in Multivariate Time Series: Towards an Automated Event Detection Method for Smart Manufacturing Systems," *Journal of Manufacturing and Materials Processing*, vol. 4, no. 3, doi: 10.3390/jmmp4030088.

[19] P. Lyu, H. Zhang, W. Yu, and C. Liu, "A novel model-independent data augmentation method for fault diagnosis in smart manufacturing," *Procedia CIRP,* vol. 107, pp. 949-954, 2022/01/01/ 2022, doi: https://doi.org/10.1016/j.procir.2022.05.090.

[20] D. Xu, Z. Zhang, and J. Shi, "A New Multi-Sensor Stream Data Augmentation Method for Imbalanced Learning in Complex Manufacturing Process," *Sensors*, vol. 22, no. 11, doi: 10.3390/s22114042.

[21] J. P. Yun, W. C. Shin, G. Koo, M. S. Kim, C. Lee, and S. J. Lee, "Automated defect inspection system for metal surfaces based on deep learning and data augmentation," *Journal of Manufacturing Systems,* vol. 55, pp. 317-324, 2020/04/01/ 2020, doi: https://doi.org/10.1016/j.jmsy.2020.03.009.

[22] Y. Wang, K. Li, S. Gan, C. Cameron, and M. Zheng, "Data Augmentation for Intelligent Manufacturing with Generative

Adversarial Framework," in *2019 1st International Conference on Industrial Artificial Intelligence (IAI)*, 23-27 July 2019 2019, pp. 1-6, doi: 10.1109/ICIAI.2019.8850773.

[23] S. R. Malakshan, M. S. E. Saadabadi, M. Mostofa, S. Soleymani, and N. M. Nasrabadi, "Joint Super-Resolution and Head Pose Estimation for Extreme Low-Resolution Faces," *IEEE Access,* vol. 11, pp. 11238-11253, 2023, doi: 10.1109/ACCESS.2023.3241606.

[24] C. Ranjan, M. Reddy, M. Mustonen, K. Paynabar, and K. Pourak, "Dataset: Rare Event Classification in Multivariate Time Series," 2019, doi: arXiv:1809.10717.

[25] C. Ranjan, *Understanding Deep Learning Application in Rare Event Prediction*. 2020.

[26] H. Patel and P. Prajapati, "Study and Analysis of Decision Tree Based Classification Algorithms," *International Journal of Computer Sciences and Engineering,* vol. 6, pp. 74-78, 10/31 2018, doi: 10.26438/ijcse/v6i10.7478.

[27] B. Charbuty and A. Abdulazeez, "Classification Based on Decision Tree Algorithm for Machine Learning," *Journal of Applied Science and Technology Trends,* vol. 2, no. 01, pp. 20 - 28, 03/24 2021, doi: 10.38094/jastt20165.

[28] M. Sheykhmousa, M. Mahdianpari, H. Ghanbari, F. Mohammadimanesh, P. Ghamisi, and S. Homayouni, "Support Vector Machine Versus Random Forest for Remote Sensing Image Classification: A Meta-Analysis and Systematic Review," *IEEE Journal of Selected Topics in Applied Earth Observations and Remote Sensing,* vol. 13, pp. 6308-6325, 2020, doi: 10.1109/JSTARS.2020.3026724.

[29] C. M. Yeşilkanat, "Spatio-temporal estimation of the daily cases of COVID-19 in worldwide using random forest machine learning algorithm," *Chaos, Solitons & Fractals,* vol. 140, p. 110210, 2020/11/01/ 2020, doi: https://doi.org/10.1016/j.chaos.2020.110210.

[30] N. P. Tigga and S. Garg, "Prediction of Type 2 Diabetes using Machine Learning Classification Methods," *Procedia Computer Science,* vol. 167, pp. 706-716, 2020/01/01/ 2020, doi: https://doi.org/10.1016/j.procs.2020.03.336.

[31] Q. Yu, J. Liu, Y. Zhang, and J. Li, "Simulation of rice biomass accumulation by an extended logistic model including influence of meteorological factors," *International Journal of Biometeorology,* vol. 46, no. 4, pp. 185-191, 2002/09/01 2002, doi: 10.1007/s00484-002-0141-3.

[32] P. Wang, X. Zheng, J. Li, and B. Zhu, "Prediction of epidemic trends in COVID-19 with logistic model and machine learning technics," *Chaos, Solitons & Fractals,* vol. 139, p. 110058, 2020/10/01/ 2020, doi: https://doi.org/10.1016/j.chaos.2020.110058.

[33] I. Ferdib Al and M. Ghosh, "An Enhanced Stroke Prediction Scheme Using SMOTE and Machine Learning Techniques," in *2021 12th International Conference on Computing Communication and Networking Technologies (ICCCNT)*, 6-8 July 2021 2021, pp. 1-6, doi: 10.1109/ICCCNT51525.2021.9579648.

[34] E. Ileberi, Y. Sun, and Z. Wang, "Performance Evaluation of Machine Learning Methods for Credit Card Fraud Detection Using SMOTE and AdaBoost," *IEEE Access,* vol. 9, pp. 165286-165294, 2021, doi: 10.1109/ACCESS.2021.3134330.

[35] B. Al Absi, M. Anbar, and S. Rihan, "Conditional Tabular Generative Adversarial Based Intrusion Detection System for Detecting Ddos and Dos Attacks on the Internet of Things Networks," *Sensors,* vol. 23, p. 5644, 06/16 2023, doi: 10.3390/s23125644.

[36] A. S. Dina, A. B. Siddique, and D. Manivannan, "Effect of Balancing Data Using Synthetic Data on the Performance of Machine Learning Classifiers for Intrusion Detection in Computer Networks," *IEEE Access,* vol. 10, pp. 96731-96747, 2022, doi: 10.1109/ACCESS.2022.3205337.

[37] S. Mangalathu, S.-H. Hwang, and J.-S. Jeon, "Failure mode and effects analysis of RC members based on machine-learning-based SHapley Additive exPlanations (SHAP) approach," *Engineering Structures,* vol. 219, p. 110927, 2020/09/15/ 2020, doi: https://doi.org/10.1016/j.engstruct.2020.110927.

[38] D.-C. Feng, W.-J. Wang, S. Mangalathu, and E. Taciroglu, "Interpretable XGBoost-SHAP Machine-Learning Model for Shear Strength Prediction of Squat RC Walls," *Journal of Structural Engineering,* vol. 147, no. 11, p. 04021173, 2021/11/01 2021, doi: 10.1061/(ASCE)ST.1943-541X.0003115.

[39] A. E. Maxwell, T. A. Warner, and L. A. Guillén, "Accuracy Assessment in Convolutional Neural Network-Based Deep Learning Remote Sensing Studies—Part 1: Literature Review," *Remote Sensing,* vol. 13, no. 13, p. 2450, 2021. [Online]. Available: https://www.mdpi.com/2072-4292/13/13/2450.

[40] A. E. Maxwell, T. A. Warner, and L. A. Guillén, "Accuracy Assessment in Convolutional Neural Network-Based Deep Learning Remote Sensing Studies—Part 2: Recommendations and Best Practices," *Remote Sensing,* vol. 13, no. 13, p. 2591, 2021. [Online]. Available: https://www.mdpi.com/2072-4292/13/13/2591.

[41] A. E. Maxwell and T. A. Warner, "Thematic Classification Accuracy Assessment with Inherently Uncertain Boundaries: An Argument for Center-Weighted Accuracy Assessment Metrics," *Remote Sensing*, vol. 12, no. 12, doi: 10.3390/rs12121905.

[42] A. Potapov, I. Colbert, K. Kreutz-Delgado, A. Cloninger, and S. Das, "PT-MMD: A Novel Statistical Framework for the Evaluation of Generative Systems," *2019 53rd Asilomar Conference on Signals, Systems, and Computers,* pp. 2219-2223, 2019.